\documentclass[preprint,review,12pt]{elsarticle}

\usepackage{amssymb}
\usepackage{amsmath}
\usepackage{lineno}
\usepackage{algorithm}
\usepackage{algpseudocode}
\usepackage{xcolor}
\usepackage{hyperref}

\pagecolor{white}

\usepackage{courier}
\newcommand{{\qlty}}{{\fontfamily{pcr}\selectfont{qlty}\space}}
\newcommand{{\qltyns}}{{\fontfamily{pcr}\selectfont{qlty}}}

\journal{Software Impacts}

\begin{document}

\begin{frontmatter}

\title{qlty: handling large tensors in scientific imaging deep-learning workflows} 

\author[label1,label2,label3]{Petrus H. Zwart}
\affiliation[label1]{organization={Center for Advanced Mathematics in Energy Research Applications, Lawrence Berkeley National Laboratory},}
\affiliation[label2]{organization={Berkeley Synchrotron Infrared Structural Biology program, Lawrence Berkeley National Laboratory},}
\affiliation[label3]{organization={Molecular Biophysics and Integrated Bioimaging Division, Lawrence Berkeley National Laboratory},
             addressline={1 Cyclotron Road},
             city={Berkeley},
             postcode={94720},
             state={California},
             country={USA}
             }

\begin{abstract}
In scientific imaging, deep learning has become a pivotal tool for image analytics. However, handling large volumetric datasets, which often exceed the memory capacity of standard GPUs, require special attention when subjected to deep learning efforts. This paper introduces \qltyns, a toolkit designed to address these challenges through tensor management techniques. \qlty offers robust methods for subsampling, cleaning, and stitching of large-scale spatial data, enabling effective training and inference even in resource-limited environments. 
\end{abstract}


\end{frontmatter}



\begin{table*}[!t]
\begin{tabular}{p{0.50\textwidth}p{0.5\textwidth}}
\hline
Current code version & 0.2.0 \\
Permanent link to code/repository used for this code version & \url{https://github.com/phzwart/qlty/releases/tag/v0.2.0} \\
Permanent link to reproducible capsule & \url{ https://codeocean.com/capsule/c5aabf63-6d0b-4e4b-abbc-99abe85870c8/} \\
Legal code license & BSD License \\
Code versioning system used & git \\
Software code languages, tools and services used & Python, Jupyter Notebook \\
Compilation requirements, operating environments and dependencies & Python, pytorch, dask, einops \\
If available, link to developer documentation/manual & \url{https://qlty.readthedocs.io/} \\
Support email for questions & \href{mailto:PHZwart@lbl.gov}{PHZwart@lbl.gov} \\
\hline
\end{tabular}
\caption{Code metadata}

\end{table*}

\section{Introduction}
\label{intro}
In computer vision applications within the experimental sciences, such as in the analysis of X-ray or Electron Tomography, Focused Ion Beam Scanning Electron Microscopy (FIB-SEM), or Adaptive Optics Lattice Light Sheet Microscopy (AO-LLSM) \cite{Le_Gros2005-yv,Lucic2005-cv,Xu2017-ql,Parlakgul2022-ln,Liu2018-bh}, the use of deep learning networks for image processing and vision-related tasks has become increasingly important. A prime example of the use of deep learning techniques is in semantic segmentation \cite{9926014,Pelt2018,Schmidt2018,Stringer2021,Lubbers2017}, where pixel-level classifications in images are made so specific objects can be extracted, or properties of a sample can be derived \cite{puma2018,puma2021}. 

When these large datasets are inspected, visualized, or analyzed, they are done as a complete object. Manual annotation often results in labeling only part of an image, the full volume is still needed to contextualize the information during the annotation process as smaller cropped sections can make proper labeling difficult. Similarly, when extracting downstream conclusions after segmentation, the spatial relationships within the full object need to remain intact. While libraries for handling large \cite{Moore2021} and very large \cite{Ruan2023.12.31.573734} data volumes are available, these methods are not directly geared towards enabling deep learning, necessitating the development of specialized tools and techniques for handling, processing, and segmenting large datasets within deep learning workflows. During the training and inference of a deep learning workflow, the associated memory requirements for large datasets rapidly outruns the available capacity conventional hardware has to offer, pushing the need for efficient data handling and processing techniques.

Out-of-core processing methods have emerged as a crucial approach to handling datasets that exceed available memory, offering a practical solution to the challenges posed by large volumetric data in scientific computer vision \cite{Ruan2023.12.31.573734}. These techniques typically involve data chunking, where the dataset is divided into manageable portions that can be loaded and processed sequentially, allowing for the analysis of data that far exceeds the capacity of RAM or GPU memory. While effective in many data processing scenarios, existing out-of-core methods often lack the specific functionality required for complex computer vision tasks, particularly those involving deep learning models. A key requirement for a tool to aid in these scenarios is the ability to handle overlapping subtensors, both during the chunking process and when stitching processed results back together. This capability is crucial for maintaining spatial continuity and addressing edge effects in convolutional neural networks. The absence of a light-weight tool that can efficiently manage patching, manipulation and stitching, while also integrating seamlessly with PyTorch's tensor operations for computer vision applications, drove the development to fill this gap. 

Here we present \qltyns, a lightweight python toolkit that addresses these challenges through a suite of specialized functions. By providing robust methods for tensor patching, manipulation, and stitching, \qlty enables researchers and developers to work effectively with large-scale spatial data, even in resource-constrained environments. We will illustrate \qltyns's functionality and demonstrate its effectiveness through a segmentation of experimental data, illustrating how \qlty can enable the processing of larger datasets that would otherwise be challenging on limited hardware.

\section{Purpose and Functionality}
\begin{figure}[h]
\centering
\includegraphics[width=\columnwidth]{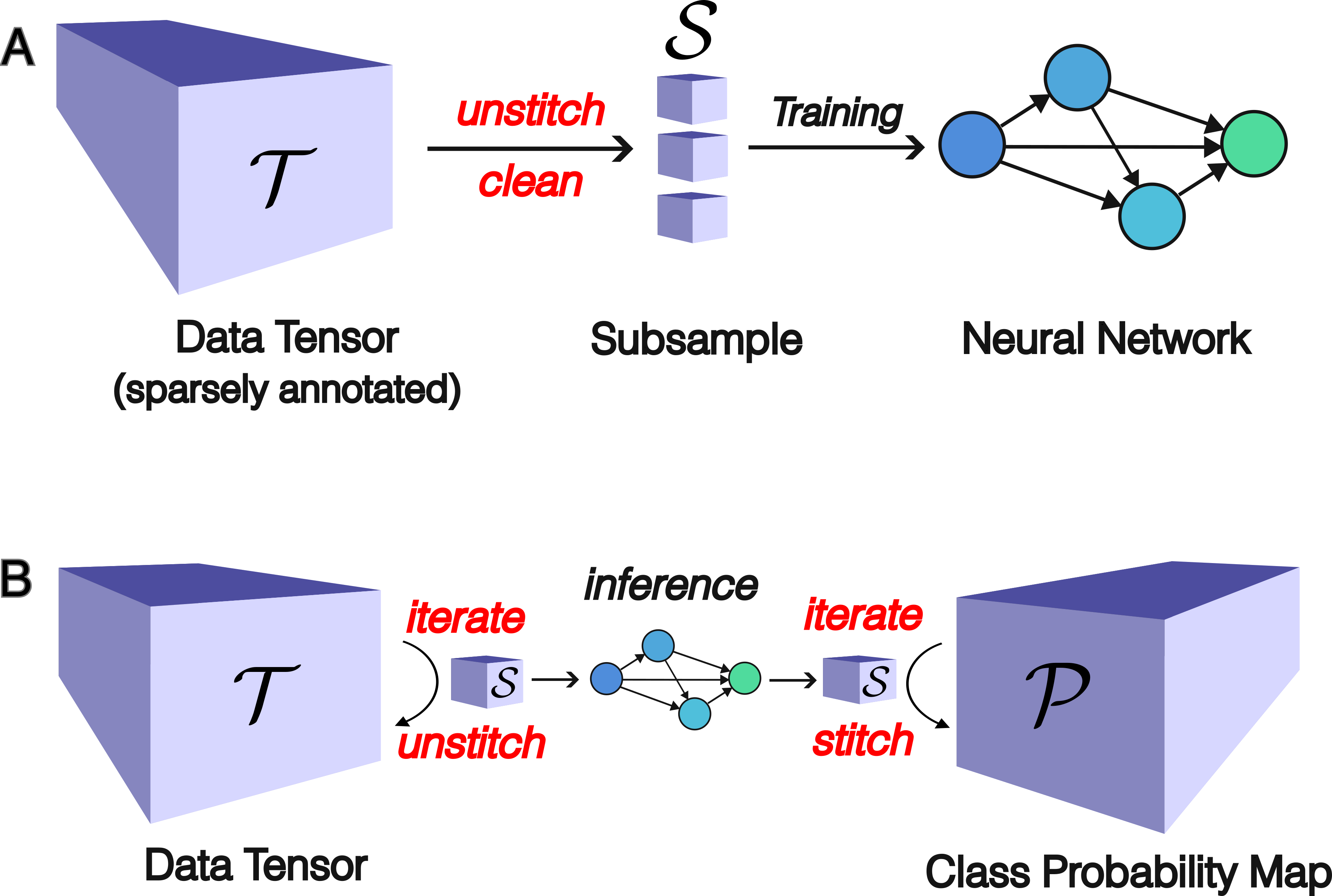}
\caption{High level overview of \qltyns's functionality, with specific methods depicted in red. A - For generating training data with a smaller spatial footprint, \qlty subsamples a sparsely annotated tensor $\mathcal{T}$ into smaller chunks $\mathcal{S}$ while discarding parts of $\mathcal{T}$ for which no training data is available. B - In inference scenarios, \qlty iterates over data tensor $\mathcal{T}$ yielding manageable chunks of data that can be subjected to a neural network of choice. The resulting output is iteratively placed back into a tensor of the right size. } \label{fig:overview}
\end{figure}

\qlty is primarily aimed at providing subsampling and out-of-core tools for segmentation, denoising and related techniques in scientific imaging. A high level overview of \qlty within these types of workflows is depicted in Figure \ref{fig:overview}

\subsection{Subsampling}
Assume we have a PyTorch tensor $\mathcal{T}$ with shape $(N,C,Y,X)$. This data structure contains $N$ items of $C$ channel data of a two-dimensional spatial map with lengths $Y$ and $X$. When subsampling $\mathcal{T}$, the aim is to obtain tensor $\mathcal{S}$ of size $(M,C,Y_w,X_w)$, where $(Y_w,X_w)$ is referred to as the \emph{window size}. The channel count $C$ remains untouched, but the resulting batch size $M$ will typically be not equal to $N$. Within \qltyns, subsampling is done in a deterministic fashion by translating a window with a predefined step size across the input tensor, while deploying an step-back correction at the edges of tensor $\mathcal{T}$ to ensure that the last subsampled window has edges aligned with the input tensor.  This process is outlined for a multi-channel 2D tensor in Algorithm \ref{alg:subsample} and extends analogously to 3D scenarios. At present, 2D + time or 3D + time scenarios are not supported, but will be added in the future.

\begin{algorithm}
\caption{Subsampling Tensor $\mathcal{T}$}
\begin{algorithmic}
\Require $\mathcal{T}$ of shape $(N, C, Y, X)$, window size $(Y_w, X_w)$, step size $(\text{step}_y, \text{step}_x)$
\Ensure Tensor with subsampled chunks $\mathcal{S}$

\State Initialize an empty list $\emph{subsamples}$
\For {$n = 0$ to $N-1$}
    \For {$y = 0$ to $Y - Y_w$ with step $\text{step}_y$}
        \For {$x = 0$ to $X - X_w$ with step $\text{step}_x$}
           \State $y_\text{start} = min(y, Y - \text{step}_y)$
           \State $y_\text{stop} = y_\text{start}+\text{step}_y$
            \State $x_\text{start} = min(x, X - \text{step}_x)$
            \State $x_\text{stop} = x_\text{start}+\text{step}_x$
            \State $\text{slice}_y = slice(y_\text{start},y_\text{stop},1)$
            \State $\text{slice}_x = slice(x_\text{start},x_\text{stop},1)$
            \State $\text{chunk} = \mathcal{T}[n:n+1,:,\text{slice}_y,\text{slice}_x]$
            \State Append $\text{chunk}$ to $\emph{subsamples}$
        \EndFor
    \EndFor
\EndFor
\State $\mathcal{S} = \text{concatenate}(\emph{subsamples}, \text{dim}=0)$
\State \Return $\mathcal{S}$
\end{algorithmic}
\label{alg:subsample}
\end{algorithm}

\subsection{Augmentation \& Cleaning}
When tensors are subsampled as described, a specific pixel or voxel from the original tensor can appear in multiple subsamples. Due to the finite size of these subsampled tensors and the receptive field of the neural network, the equivalence of identical observations across subsamples is disrupted. This issue is most evident for pixels at the edges of the subsampled tensors, which, at best, only have half the number of unique neighbors within their vicinity during convolutional calculations. When working with semantic segmentation data, a border region in the subsampled tensor can be defined, which allows us to mark data as \emph{unobserved} and exclude these labels from the training dataset, only retaining internal voxels or pixels. 

While the above procedure introduces a sparsification of the training data by excluding edge pixels or edge voxels, it is far more common that the sparsity of labeled data originates from selected human labeling. In such cases, a large amount of subsampled tensors can contain no training data due to absence of labels. \qlty provides tools to remove or exclude such tensor pairs, enriching the information content of the data, while ensuring it runs on the available hardware.  

\subsection{Stitching}
Even though the memory requirements for inference are less than those for training, subsampling can still be necessary. In \qltyns, the same subsampling algorithm to generate data for inference is used as in the training scenario, without the cleaning step. In addition, accumulator arrays are created,  a mean array and a normalization array, where the inference results of subsampled tensors are stored. In the mean array, inference results are multiplied by a weight term that accounts for the relative significance of predictions for border pixels and added to the accumulator. The normalization array accumulates the sum of these weights. The final prediction is obtained by calculating the weighted mean, achieved by dividing the mean array by the normalization array. Both the mean and accumulator arrays are \emph{zarr} arrays, caching the results on disc. The final normalization is performed using \emph{dask}, making use of multiprocessing options when available.

\section{Example Usage}
The use of \qlty in a segmentation setting is illustrated by a segmentation of a 3D cryo-electron tomographic reconstruction of the cell derived matrix (CDM) produced by human telomerase immortalized foreskin fibroblasts (TIFFS) \cite{Zens2024-cj}, publicly available via EMPIAR \cite{Iudin2023-ge}. For illustrative purposes, a minimal manual annotation was conducted using napari \cite{napari2019}, where 52k pixels \-- the equivalent of 5 data cubes of length 64 \-- were labeled in a few slices across the XY,YZ and XZ projections. \qlty was run using typical settings, with a target window of $(64,64,64)$. Due to the low number of labeled pixels, a step size of $32$ in all directions was used, resulting in duplicating the number of annotated pixels by a factor of 7.46, but reducing the total data volume to analyze by a factor of 9 by a subsequent elimination of subtensors without annotated data. Additional details are found in Table \ref{tab:train}.  

\begin{table}[h!]
\label{tab:train}
\centering
\begin{tabular}{ll}
\hline
\textbf{Training Scenario Settings}&\\
\hline
Full Tensor shape& $(1,1,236,720,510)$\\
Annotated  \emph{Non Filament} voxels &   51505\\
Annotated  \emph{Filament} voxels &   1809 \\
Window shape & $(64,64,64)$ \\
Step Size & $32$ in $Z$,$Y$,$X$\\
Border Size & $0$ in  $Z$,$Y$,$X$\\
Total subsample &  $(2310,1,64,64,64)$ \\
Clean subsample & $(256,1,64,64,64)$ \\
Data duplication rate & 7.47 \\
\hline
\end{tabular}
\caption{\qlty training settings. Sparse labeling of a tensor, approximately 0.06\% of voxels were annotated , resulted in 256 cube-like tensors of length 64. The translating window approach results in annotated data duplication - identical voxels appearing in multiple subsamples - at a rate of 7.47.}
\end{table}

The generated training data was used to train an ensemble of eight Sparse Mixed Scale neural Networks (SMSNets) with random architecture as available from the \emph{DLSIA} library \cite{Roberts:yr5117}. Using Adam as an optimizer and a cross-entropy target with a learning rate of 0.01 and batch size of 16,  fifty epochs on an A100 GPU yielded an average macro F1 score \-- taking into account the class imbalance \-- of  89\% and 88\% for the training and validation data respectively, across all 8 models. 

These trained networks were used to run inference across the full data volume. For inference, a larger window size is technically feasible because of lower memory requirements. Using a window size of 128, a step size of 108 and a border width of 10 were used. By setting the border weight to zero, the resulting stitching operation results in patching together non-overlapping tensor of size $(108,108,108)$. In this manner, 231 subtensors were generated and run through an inference pass. Further \qlty setting are found in Table \ref{tab:inference}.      
\begin{table}[h!]
\label{tab:inference}
\centering
\begin{tabular}{ll}
\hline
\textbf{Inference Scenario Settings}&\\
\hline
Input tensor shape& $(1, 1, 236,720,510)$\\
Window shape & $(128,128,128)$ \\
Step size & $108$ in $Z$,$Y$,$X$\\
Border size & $10$ in  $Z$,$Y$,$X$\\
Border weight & $0$ in  $Z$,$Y$,$X$\\ 
Subsampled tensor shape & $(231,1,128,128,128)$ \\ 
Output tensor shape & (1, 2, 236,720,510) \\
\hline
\end{tabular}
\caption{\qlty inference settings. Using the ensemble of trained network, inference was performed on subsampled tensors of with spatial footprint of $(128,128,128)$, with a step size of 108 and a border size of 10. After stitching the subtensors, a tensor of shape (1,2,236,720,510) was obtained.}
\end{table}

\begin{figure}[h]
\centering
\includegraphics[width=0.5\columnwidth]{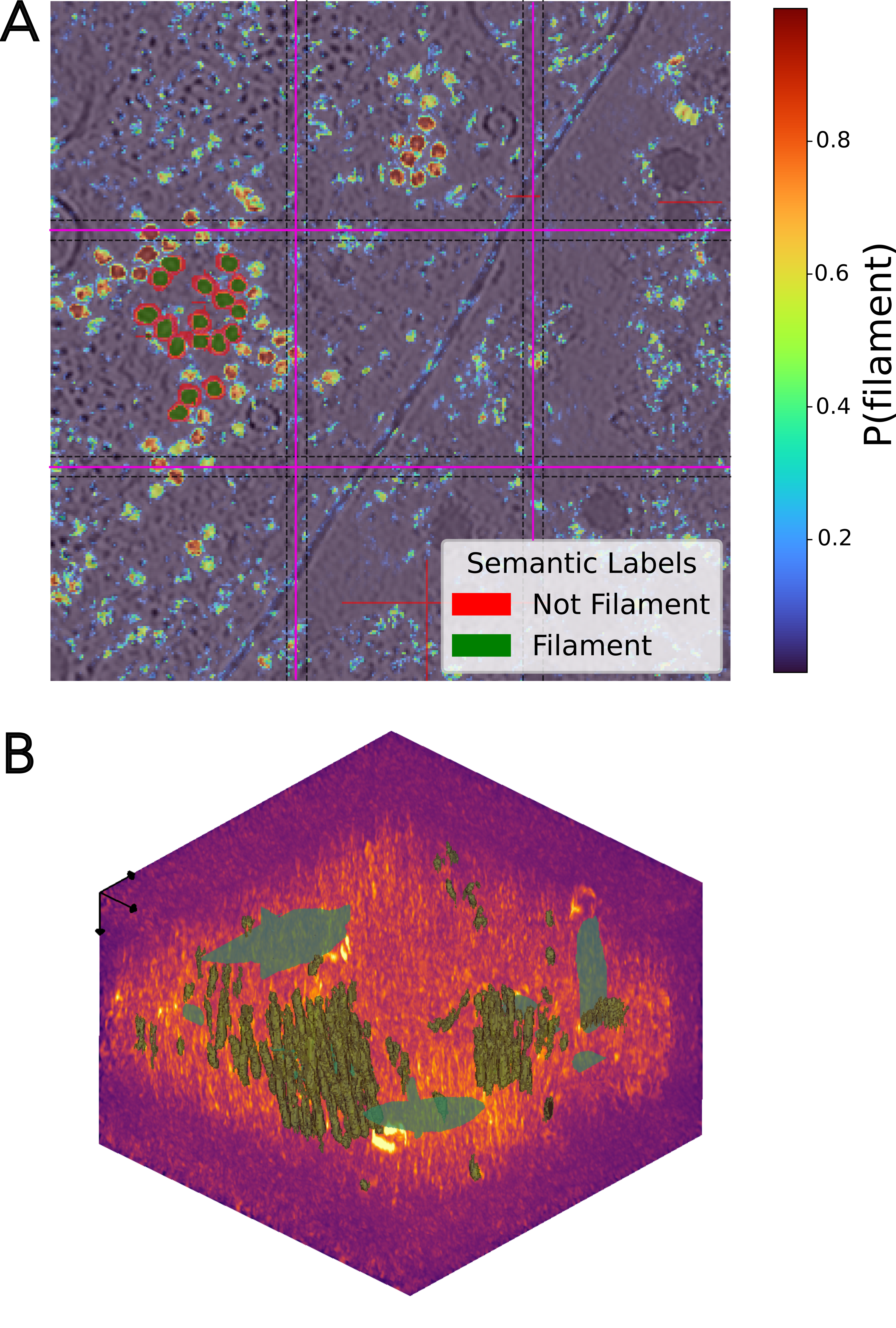}
\caption{Segmentation results. A \-- A 2D section of the resulting segmentation of the tomographic data (grey) is overlaid with the class probability map for the intermediary filaments (multi-color heatmap) and part of the hand annotated voxels (green, red). Boundaries of the subsampled tensor are shown as black doted lines, while the purple solid line indicates the location of the seam where two adjacent tensors meet when taking into account the border weighting procedure. B \-- A 3D overview of the tomographic density (magma heatmap), hand-labeled voxels (light-green 2D patches) and resulting class labels in 3D for the filaments from the probability map after a simple connected component sizing filter (bright green).} \label{fig:results}
\end{figure}

The results of the inference are shown in Figure \ref{fig:results}. The high quality of the depicted segmentation originates in large part from the data augmentation by the windowed subsampling of the tensor and the associated data duplication that follows from this procedure. 

\section{Impact}
The \qlty toolkit offers a light-weight solution for handling large-scale (volumetric) data for computer vision tasks in scientific imaging. By addressing the challenges associated with processing large datasets that exceed the memory capacity of typical GPUs, \qlty enables efficient and effective image analysis in resource-constrained environments. This capability is crucial for a variety of scientific imaging applications in biology and material sciences.

\qlty has been put to practical use in a number of applications, such as the segmentation of contaminant peaks in X-ray powder diffraction \cite{yanxon2023imagesegmentationusingunet}, the segmentation of fibers from X-ray tomography scans of reinforced concrete \cite{Roberts:yr5117} and near-real time segmentation of X-ray tomography datasets at large scale facilities \cite{als_machine_learning_2024}. In a denoising setting, \qlty has allowed for routine training and inference of X-ray tomographic data exceeding $10^9$ pixels on a RTX-3090 NVIDIA consumer hardware GPU \cite{qafoku_denoising_2024}. Training and inference on this type of data without chunking would require at least an order of magnitude more memory than the 24GB that GPU possesses, making \qlty a indispensable tool in these workflows. 

The practical impact of \qlty as observed in the aforementioned applications \cite{Roberts:yr5117,yanxon2023imagesegmentationusingunet,als_machine_learning_2024}, lies in its integration within standard workflows, allowing users or automated processes to adjust window and step sizes to easily manage GPU memory limitations. It is commonly embedded as a preprocessing step in segmentation or denoising pipelines, simplifying the process of running these workflows across different hardware architectures, such as RTX 3090s or A100 GPUs. While memory constraints are often addressed by adjusting batch sizes, breaking down tensors into smaller subtensors to fit within these limitations is typically more complex — \qlty effectively streamlines this process.

The seamless integration of \qlty within the PyTorch framework, streamlines machine learning workflows for scientific imaging. \qltyns’s subsampling and stitching functions can be easily incorporated within existing data analytics pipelines \cite{Roberts:yr5117}, circumventing the need to write \emph{ad-hoc} functionality that addresses these issues. \qlty thus reduces a barrier to utilize advanced image processing techniques in the broader scientific community.

\section{Future Directions}
By working closely with experimental scientists, a number of future development directions have been identified.

\emph{Supporting Fragmented Workflows:} In many applications, the locations where data is generated, computation is performed, and where results are visualized are often different and distributed across various network locations. A promising future direction for \qlty is to integrate it with technologies that allow seamless data access across these disparate networked environments. By combining \qlty with distributed computing frameworks and remote data access protocols, we can facilitate efficient data handling and computation even in workflows fragmented across many locations, ensuring smooth operations regardless of where the data resides.

\emph{Supporting Image Pyramids:} Image pyramids are powerful data structures for managing large datasets across different scales. Neural networks that leverage these pyramids can effectively decouple receptive field requirements across scale spaces, potentially leading to more efficient models. A key future direction is to extend \qlty to independently partition large tensors within these scale spaces. By enabling the processing of image data across various scales, and combining the resulting outputs only at the end, \qlty can help create leaner and more effective neural networks for large-scale imaging tasks.

\emph{Tracking Positional Embeddings:}  Aligning large images across modalities poses significant challenges, and the use of positional information is crucial in such tasks. Currently, tracking positional embeddings across subtensors can be achieved by creating a dummy tensor with positional data. However, a more efficient approach would be to enhance \qlty by attaching origin and slicing parameters to each chunk. These parameters could then be used to maintain positional context and enable tensor-index to location parameters in a physical model throughout the processing pipeline, improving the accuracy and effectiveness of alignment operations.

\section{Acknowledgements}
We gratefully acknowledge the support of this work by the Laboratory Directed Research and Development Program of Lawrence Berkeley National Laboratory under US Department of Energy contract No. DE-AC02-05CH11231. Further support originated from the Center for Advanced Mathematics in Energy Research Applications funded via the Advanced Scientific Computing Research and the Basic Energy Sciences programs, which are supported by the Office of Science of the US Department of Energy (DOE) under contract No. DE-AC02-05CH11231, and from the National Institute of General Medical Sciences of the National Institutes of Health (NIH) under award 5R21GM129649-02. We acknowledge the use of resources of the National Energy Research Scientific Computing Center (NERSC), a U.S. Department of Energy Office of Science User Facility operated under Contract No. DE-AC02-05CH11231, under NERSC allocation m4055.
During the preparation of this work the author used chatGPT in order to correct grammar and improve sentence construction. After using this tool/service, the author reviewed and edited the content as needed and takes full responsibility for the content of the publication.

\bibliographystyle{unsrt}
\bibliography{qlty.bib}

\end{document}